\documentclass[conference]{IEEEtran}
\ifCLASSINFOpdf
  \usepackage[pdftex]{graphicx}
  \DeclareGraphicsExtensions{.pdf,.jpeg,.png}
\else
\fi
\usepackage{amsmath}
\usepackage{array}
\usepackage{url}

\usepackage{xspace}
\usepackage{booktabs}
\usepackage[keeplastbox]{flushend}

\newcommand{\swat}{SWaT\xspace}
\newcommand{\cps}{CPS\xspace}
\newcommand{\plc}{PLC\xspace}
\newcommand{\plcs}{PLCs\xspace}

\begin{document}
\IEEEoverridecommandlockouts
%
\title{Anomaly Detection for a Water Treatment System Using Unsupervised Machine Learning}



%
\author{\IEEEauthorblockN{ Anonymous }
\IEEEauthorblockA{\IEEEauthorrefmark{1}
An organization\\
Somewhere\\ Email: mail@example.com}
\IEEEauthorblockA{\IEEEauthorrefmark{2}Another organization\\
Somewhere\\ Email: mail@example.com}}
\author{\IEEEauthorblockN{Jun Inoue\IEEEauthorrefmark{1},
Yoriyuki Yamagata\IEEEauthorrefmark{1},
Yuqi Chen\IEEEauthorrefmark{2}\thanks{This work was supported in part by the National Research Foundation~(NRF), Prime Minister's Office, Singapore, under its National Cybersecurity R\&D Programme (Award No.~NRF2014NCR-NCR001-040) and administered by the National Cybersecurity R\&D Directorate.},
Christopher M. Poskitt\IEEEauthorrefmark{2} and
Jun Sun\IEEEauthorrefmark{2}}
\IEEEauthorblockA{\IEEEauthorrefmark{1}
National Institute of Advanced Industrial Science and Technology (AIST)\\
Ikeda, Japan\\ Email: \{jun.inoue, yoriyuki.yamagata\}@aist.go.jp}
\IEEEauthorblockA{\IEEEauthorrefmark{2}Singapore University of Technology and Design\\
Singapore, Singapore\\ Email: yuqi\_chen@mymail.sutd.edu.sg; \{chris\_poskitt, sunjun\}@sutd.edu.sg}}


\maketitle
\begin{abstract}
In this paper, we propose and evaluate the application of \emph{unsupervised machine learning} to anomaly detection for a \emph{Cyber-Physical System (CPS)}.
We compare two methods: \emph{Deep Neural Networks (DNN)} adapted to time series data generated by a \cps, and \emph{one-class Support Vector Machines (SVM)}.
These methods are evaluated against data from the \emph{Secure Water Treatment (SWaT)} testbed, a scaled-down but fully operational raw water purification plant.
For both methods, we first train detectors using a log generated by SWaT operating under normal conditions.
Then, we evaluate the performance of both methods using a log generated by SWaT operating under 36 different attack scenarios.
We find that our DNN generates fewer false positives than our one-class SVM while our SVM detects slightly more anomalies.
Overall, our DNN has a slightly better F measure than our SVM. 
We discuss the characteristics of the DNN and one-class SVM used in this experiment, and compare the advantages and disadvantages of the two methods.
\end{abstract}


%
\IEEEpeerreviewmaketitle

\section{Introduction}

A \emph{Cyber-Physical System (CPS)} is a complex system consisting of distributed computing elements that interact with physical processes.
CPSs have become ubiquitous in modern life, with software now controlling cars, airplanes, and even critical public infrastructure such as water treatment plants, smart grids, and railways.

\emph{Anomaly detection} for CPSs concerns the identification of unusual behaviors (anomalies), i.e.~behaviors that are not exhibited under normal operation.
These anomalies may result from attacks on the control elements, network, or physical environment, but they may also result from faults, operator errors, or even just standard bugs or misconfigurations in the software.
The ability to detect anomalies thus serves as a defensive mechanism, while also facilitating development, maintenance, and repairs of CPSs.

Anomaly detection techniques can be rule-based or model-based.
In the former, rules are supplied that capture patterns in the data, and detection involves testing for their violation.
In the latter, a mathematical model characterizing the system is supplied, and detection involves querying new data against that model (e.g.~\cite{Pasqualetti2011,Teixeira2012,Verma2004,Narasimhan2007,Hofbaur2002,Zhao2005,Hofbaur2004}).
Unfortunately, constructing models of CPSs that are accurate enough in practice is a notoriously difficult task, arising from the tight integration of algorithmic control and complex physical processes.

In this paper, we investigate the application of \emph{unsupervised machine learning} to building models of CPSs for anomaly detection.
The advantage of unsupervised machine learning is that it does not require any understanding of the complexities of the target CPS; instead, it builds models solely from data logs that are ordinarily available from historians.
This research direction is seeing increasing interest (e.g.~\cite{Jones2014,harada2017log}), but much remains to be understood about how to apply it effectively in practice.

We apply and compare two unsupervised methods: first, a \emph{Deep Neural Network (DNN)} consisting of a layer of \emph{Long Short-Term Memory (LSTM)} architecture followed by feed-forward layers of multiple inputs and outputs; second, a \emph{one-class Support Vector Machine (SVM)}~\cite{scholkopf2001estimating}, which is widely used for anomaly detection.

We evaluate these methods using logs from \emph{Secure Water Treatment (SWaT)}, a testbed built at the Singapore University of Technology and Design for cyber-security research~\cite{SWaT-Reference}.
SWaT is a scaled-down but fully operational water treatment plant, capable of producing five gallons of drinking water per minute.
For learning our models, we make use of a real log generated by SWaT over seven days of continuous operation, and to evaluate them, we use another four days of logged data during which the system was subjected to 36 different network attack scenarios~\cite{Goh-et_al16a}.
We compare how many faults (i.e.~anomalies) the two methods can find.

According to Aggarwal~\cite{Aggarwal:2015:OA:2842756}, there are two approaches for finding outliers (i.e.~anomalies) in time series data.
The first approach is to find outlier \emph{instants} in the time series, typically based on the deviation from predicted values at each time instant.
Many models for predicting values are proposed, such as auto-regressive models and hidden variable-based models.
However, these methods assume linearity of the system.
SWaT is a hybrid system integrating non-linear dynamical systems with digital control logic, hence these methods are unsuitable for our task.
We choose DNNs as a prediction model because DNNs can learn non-linear relations without any prior knowledge of the system.
Using an LSTM architecture, we can capture the dynamic nature of SWaT.

The second approach is to find unusual shapes in the time series.
We use one-class SVMs for this purpose.
Clustering-based methods would be difficult to use because of the high dimensionality of the data stream: the SWaT testbed contains 25 sensors and 26 actuators.

We find that the DNN performs slightly better than the one-class SVM: the precision of the DNN is higher, while recall is slightly better with the SVM.
This difference is largely accounted for by the tendency of one-class SVM to report abrupt changes in sensor values as anomalies, even when they are normal, thus causing it to report more false positives.
Our DNN and SVM usually detect anomalies when a sensor reports a constant anomalous value.
However, both methods have difficulties in detecting a gradual anomalous change of sensor values or anomalous actuator movements.
Our DNN also has difficulties in detecting out of bound values.
In general, SVM is more sensitive to anomalies---but not always.

This paper is organized as follows.
Section \ref{sec:related_works} summarizes related work and clarifies our contributions.
Section \ref{sec:swat_testbed} introduces the SWaT testbed.
Section \ref{sec:methods} introduces our DNN- and SVM-based detection methods.
Section \ref{sec:implementation} describes how our methods are implemented and our experimental setup.
We also discuss the computation costs of both methods.
Section \ref{sec:tune} describes how we tune the hyper-parameters of the methods.
Section \ref{sec:evaluation} describes our findings based on our evaluation using a SWaT attack log.
Section \ref{sec:concl} presents conclusions and future work.
In particular, we discuss the advantages and disadvantages of the two methods.

\section{Related Work}\label{sec:related_works}
There is a large body of work on simulation and model-based anomaly detection for CPSs, e.g.~\cite{Pasqualetti2011,Teixeira2012,Verma2004,Narasimhan2007,Hofbaur2002,Zhao2005,Hofbaur2004}.
However, these approaches require prior knowledge of the system's configuration, in addition to operation logs.

There is a proposal to use \emph{supervised machine learning}~\cite{Chen} to obtain a model for anomaly detection, which requires access to the source code of the control elements.
In this approach, the detector is trained on correct and incorrect behaviors, the latter of which are generated by randomly injecting faults into the control code.
Currently, only a preliminary investigation of the approach has been performed.

Jones et~al.~\cite{Jones2014} propose an SVM-like algorithm which finds a description in a \emph{Signal Temporal Logic (STL)} formula of the known region of behaviors.
An advantage of this approach is that it often creates a readable description of the known behaviors.
However, if the system behavior does not allow for a short description in STL, this method will not work.
Because SWaT is dynamic, non-linear, stochastic, and has high dimensionality, a short description is unlikely.
Moreover, in their method, the tightness function is heuristic and no justification is given.

Anomaly detection, beyond the specific application to CPSs, is a well-studied area of research (see e.g.~the survey~\cite{Chandola2009} and textbook~\cite{Aggarwal:2015:OA:2842756}).
Harada et~al.~\cite{harada2017log} applies one of the most widely-used anomaly detection methods, \emph{Local Outlier Factor (LOF)}~\cite{Breunig2000}, to an automated aquarium management system and detects the failure of mutual exclusion.
However, LOF is a method to find outliers without prior knowledge of the normal behaviors.
Because the normal behaviors are known in our case, LOF is not suitable for our task.

There is some work applying DNNs for anomaly detection.
Malhotra et~al.~\cite{Malhotra2015} use stacked LSTMs and prediction errors for detecting anomalies.
However, prediction errors for data that represents normal situations can fluctuate depending on the system state, hence using a uniform threshold to detect anomalies may not produce the best performance.
In fact, recall for their method is in the order of 10\%.
In contrast, our DNNs directly compute the probability distribution of the next status for \emph{every} time step, thus avoiding this difficulty.
In addition, our method can simultaneously handle a mixture of discrete valued data and real-number valued data, for which the prediction error is difficult to define.
Zhai et~al.~\cite{Zhai2016} propose the use of energy-based models and energy functions for detecting anomalies.
However, their work assumes that the data is real-number valued, and it is not clear whether their method can be extended to data which is a mixture of discrete and real. 
Furthermore, their energy function seems to compute energy for the entire time series, and it is also not clear how their method can be used to detect the time instants of anomalies.

The SWaT testbed and its dataset~\cite{Goh-et_al16a} have been used to evaluate a number of other approaches for cyber-attack prevention, including learning classifiers from data~\cite{Chen,Goh_et-al17a}, monitoring network traffic~\cite{Ghaeini-Tippenhauer16a}, or monitoring process invariants~\cite{Adepu-Mathur16a,Adepu-Mathur16b}.
These process invariants are derived from the physical laws concerning different aspects of the SWaT system, and thus in our terminology can be considered in the category of rule-based anomaly detection methods.

Goh et~al.~\cite{Goh_et-al17a} propose a similar unsupervised machine learning approach to learn a model of SWaT.
They use stacked LSTMs to detect anomalies, and the same SWaT dataset in their evaluation~\cite{Goh-et_al16a}.
However, they only apply their approach to the first SWaT subsystem (of six), and consider only ten attacks in their evaluation (i.e.~the attacks targeted to that subsystem).
Their anomaly detection is based on cumulative sums of prediction error for each sensor, with the evaluation based upon the number of attacks detected.
Nine of the ten attacks are detected, with four false positives reported.
In contrast, we apply our method to the SWaT testbed in its entirety (i.e.~all six subsystems) and evaluate against the full attack log, spanning 36 attacks.
We achieve very high precision, i.e.~very few false positives while a moderate recall rate.
Precision and recall are calculated based on the number of detected log entries instead of number of attacks, which should lead to smaller recall rates.
Our methods are based on probabilistic density estimation, rather than prediction error.

\section{Secure Water Treatment (SWaT) Testbed}\label{sec:swat_testbed}

\begin{figure}[!t]
	\centering
	\includegraphics[width=\linewidth]{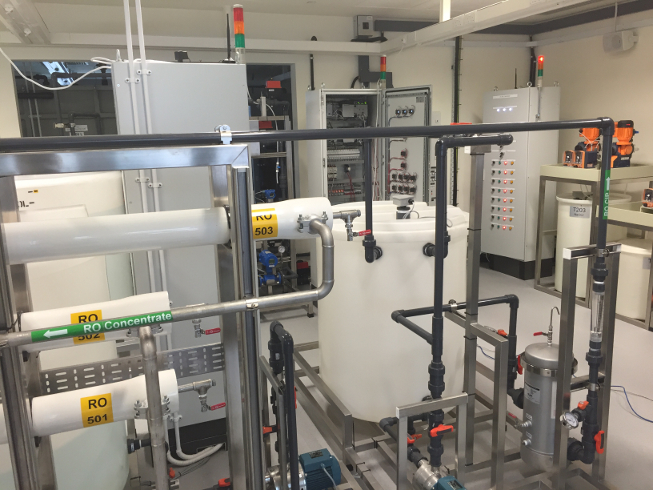}
	\caption{The Secure Water Treatment (\swat) testbed}
	\label{fig:swat_testbed}
\end{figure}

The \cps we evaluate our learning architecture on is \emph{Secure Water Treatment (\swat)}~\cite{SWaT-Reference}, a testbed at the Singapore University of Technology and Design that was built to facilitate cyber-security research (Fig.~\ref{fig:swat_testbed}).
\swat is a scaled-down but otherwise fully operational raw water purification plant, capable of producing five gallons of safe drinking water per minute.
It is representative of CPSs typically used in public infrastructure, and is highly dynamic, with the flow of water and chemicals between tanks an intrinsic part of its operation and behavior.
Raw water is treated in a modern six-stage architecture, consisting of physical processes such as ultrafiltration, de-chlorination, and reverse osmosis.
The cyber part of \swat consists of Programmable Logic Controllers (\plcs), a layered communications network, Human-Machine Interfaces (HMIs), a Supervisory Control and Data Acquisition (SCADA) workstation, and a historian.

Each stage of \swat is controlled by a dedicated \plc, which interacts with the physical environment via sensors and actuators connected over a ring network.
While varying from stage-to-stage, a typical sensor in \swat might read the level of a water tank or the rate of water flow in a pipe, and a typical actuator might operate a motorized valve for opening or closing an inflow pipe.
Each \plc repeatedly cycles through a ladder logic program, reading the latest data from the sensors and computing the appropriate signals to send to the actuators.
This data is also made available to the SCADA system and is recorded by the historian, allowing for offline analyses~\cite{Goh-et_al16a}.

Many of the security concerns associated with the automation of public infrastructure are exemplified by \swat.
If an attacker is able to compromise its network or \plc programs, they may be able to drive the system into states that cause physical damage, e.g.~overflowing a tank, pumping an empty one, or mixing chemicals unsafely.
\swat has thus been used by researchers as a testbed for developing different attack prevention measures for water treatment plants, e.g.~by monitoring process invariants~\cite{Adepu-Mathur16a,Adepu-Mathur16b}, monitoring network traffic~\cite{Ghaeini-Tippenhauer16a}, or learning and monitoring log-based classifiers~\cite{Chen,Goh_et-al17a}.

For evaluating our learning architecture, we make use of a large dataset that was obtained from the \swat historian~\cite{Goh-et_al16a}.
The dataset (available online~\cite{SWaT-Dataset}) consists of all network traffic, sensor data, and actuator data that was systematically collected over 11 days of continuous operation.
For seven of the days, \swat was operated under normal conditions, while on the other four days, it was subjected to 36 attack scenarios~\cite{Goh-et_al16a} representative of typical network-based attacks on CPSs.
These attacks were implemented through the data communications link of the \swat network: data packets were hijacked, allowing for sensor data and actuator signals to be manipulated before reaching the PLCs, pumps, and valves (e.g. preventing a valve from opening, or reporting a water tank level as lower than it actually is).
The attacks were systematically generated with respect to the sensors and components. Of the 36 attacks, 26 are single-state single-point attacks, with the remaining ones more complex: 4 are single-stage multi-point; 2 are multi-stage single-point; and 4 are multi-stage multi-point.
A full description of the attacks is provided with the dataset~\cite{SWaT-Dataset}.

\section{Anomaly Detection Methods}\label{sec:methods}
We present our two anomaly detection methods, based respectively on a DNN and one-class SVM.

\subsection{Deep Neural Network}
The first of our two anomaly detection methods uses a DNN to implement \emph{probabilistic outlier detection}.
Such a detection method requires a probability distribution for data, in which data points assigned a low probability are judged as \emph{outliers} (e.g.~data points generated by a different mechanism from normal data points).

In our method, this probability distribution is represented by a DNN that was trained on normal data from the CPS log.
Assume we have $n$ actuators and $m$ sensors.
The log under consideration is a sequence of log entries $\langle a_1, \ldots, a_n, s_1, \ldots, s_m \rangle$.
We abbreviate $a_1, \ldots, a_n$ to $\overline{a}$ and $s_1, \ldots, s_m$ to $\overline{s}$.
Further, the $i$-th log entry is denoted by $l_i \equiv \langle \overline{a}(i), \overline{s}(i) \rangle$ and the log entries up to $i$ are denoted by $\mathbf{l_{i}}$.
Our DNN computes an \emph{outlier factor},
$- \log q(l_i \mid \mathbf{l}_{i-1})$,
where $q$ is a probability distribution.
However, it cannot be computed directly in this form because there are several combinations of actuator positions, and infinitely many sensor values.
Instead, we decompose the outlier factor into:
\begin{multline}
  - \sum_{j = 1}^n \log q_j(a_j(i) \mid \mathbf{l}_{i-1}, a_1(i), \ldots, a_{j-1}(i)) \\
  - \sum_{k = 1}^m \log r_k(s_k(i) \mid \mathbf{l}_{i-1}, \overline{a}(i), s_1(i), \ldots, s_{k-1}(i))
\end{multline}
Here, $q_j$ is a discrete distribution, whereas $r_k$ is approximated by a Gaussian distribution.
We represent $q_j$ and $r_k$ using a single neural network with multiple inputs and outputs.

\begin{figure}[!t]
\centering
\includegraphics[width=0.8\linewidth]{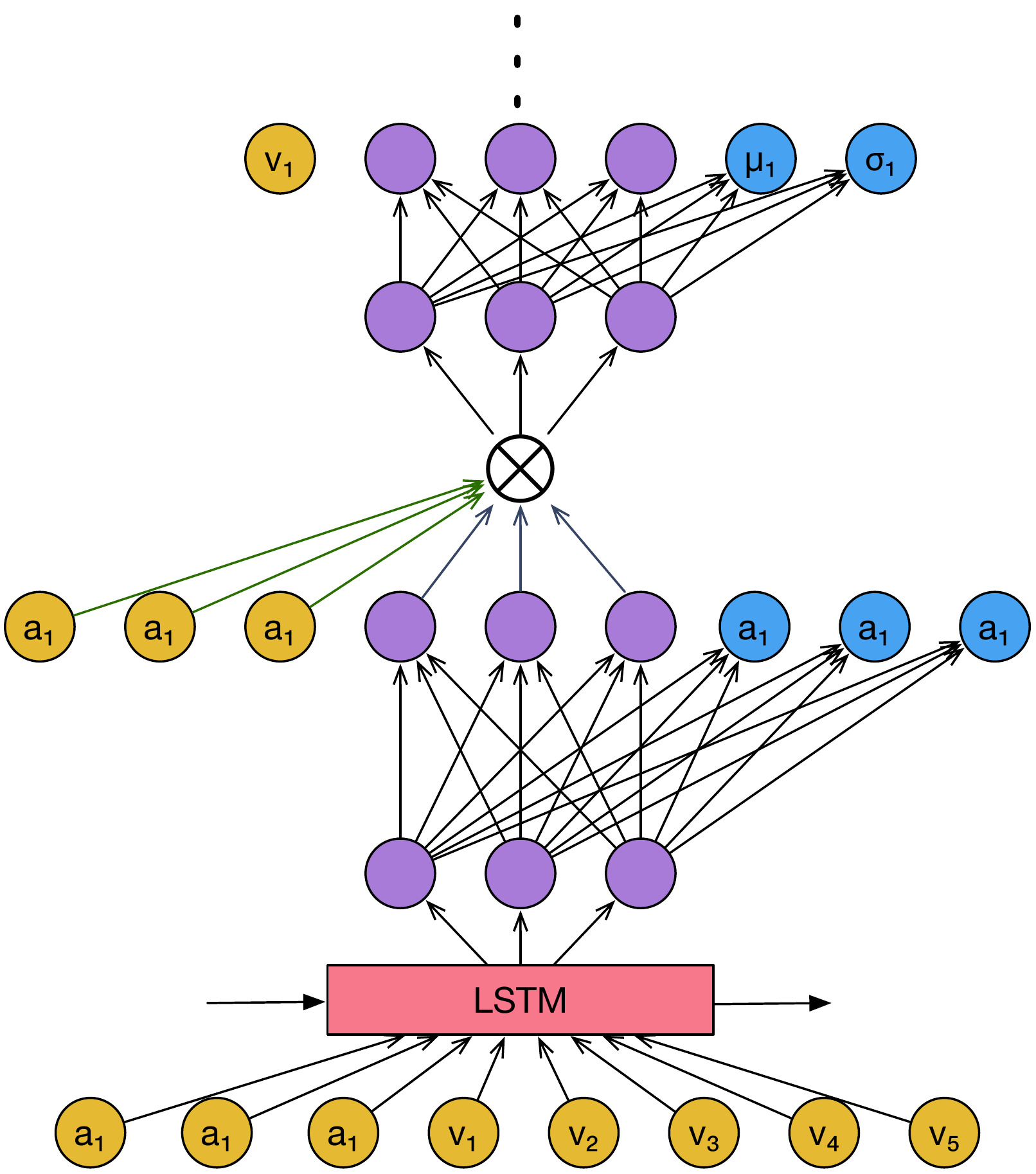}
\caption{DNN architecture}
\label{fig:DNN}
\end{figure}

Ultimately, we obtain an architecture similar to the illustration in Fig.~\ref{fig:DNN}.
The figure assumes that we have only one actuator (with three positions) and five sensors.
To compute the $i$-th outlier factor, our DNN takes log entries up to the \mbox{$i-1$-th} position into an LSTM.
The output of the LSTM is first used for predicting the probability of each actuator position using the soft-max function.
The output layer then takes the true actuator position in the $i$-th log entry and mixes it with the value of the hidden layer using a bilinear function.
The output of the bilinear function is then fully connected to the next hidden layer and output.
The outputs are the predicted mean and variance of the first sensor value.
Furthermore, the output layer takes the true first sensor value in the $i$-th log entry, and repeats the process above.
Finally, we sum up all the log probabilities of the actuator positions and sensor values to obtain the outlier factor.

To train the neural network, we must define a cost function that calculates the cost (error) of the predicted probability distributions in comparison to the observed log entries.
We use the cross entropy of the real probability distributions $p$ and the predicted probability distributions $q$ of the log entries.
The cross entropy $C(p, q)$ between $p$ and $q$ is defined as follows:
\begin{equation}
  C(p, q) = \int - p(z) \log q(z) d\mu
\end{equation}
where $z$ is a data point and $\mu$ is an appropriate measure over the data points.
However, we do not know the true distribution, $p$; instead, we approximate it by Dirac's delta function, $\delta(z - x)$, where $x$ is an observation.
Therefore, we minimize:
\begin{align}
  C(p, q) &= - \log q(x)\\
          &= \sum_{i = 1}^N - \log q(l_i \mid \mathbf{l}_{i-1})
\end{align}
This is the sum of all the outlier factors.
Thus, we use the sum of all the outlier factors in the training data as the cost function.

Once we have trained a DNN, we can compute outlier factors as stated above.
To determine outliers, however, we need to have a threshold for outlier factors.
This is difficult to do theoretically, so we need to use experiments.

\subsection{One-Class SVM}

Our experiments compare our DNN-based method with a straightforward application of the widely used one-class SVM algorithm~\cite{scholkopf2001estimating}.
To learn a non-linear classification boundary, we use the \emph{Radial Basis Function (RBF)} kernel.

Because the log sequence is a time series, we employ the sliding window method~\cite{Dietterich2002} to convert the data into individual feature vectors.
If $l_i$ denotes the $i$-th log entry and $w$ is a prescribed window size, then a tuple of the form $W_i \equiv \langle l_i, l_{i+1}, \ldots, l_{i+w-1}\rangle$ is called a \emph{window}, and the one-class SVM classifies each window as normal or abnormal.
A fixed-size window is slid across the entire log, so that from a log with $k$ entries, $k-w+1$ windows $W_1, W_2, \ldots, W_{k-w+1}$ are extracted and classified.

In the training phase, we simply extract all windows from the training data and feed them to the training algorithm of one-class SVM.
In the testing phase, we use test data that has a normal/abnormal label for every log entry.
We extract windows as we did for training data, and a label is derived for each window from the log entries they contain as follows: a window $W_i$ is labeled abnormal if at least one of the log entries $l_i, l_{i+1}, \ldots, l_{i+w-1}$ is labeled abnormal; otherwise it is labeled normal.
Each window is fed to the trained classifier, which outputs a normal/abnormal verdict; this verdict is compared to the label of the window for evaluation.

Essentially, we evaluate the classifier by its ability to pick up anomalies occurring in a window regardless of the location.
In graphical presentations of experimental results, we align the verdict of one-class SVM with the beginning of the window that it is associated to.
Thus, technically, the first occurrence of the truly anomalous log entry may be off by up to $w-1$ entries in the graphs.
In this sense, the resolution of SVM's verdicts degrades as the window size is increased.
In the cases we have tried, however, the resolution is too fine to affect our conclusions.

We normalize data based on means and variances of each (training / testing) dataset.
A preliminary logarithmic grid search found that the specific method of normalization has a large effect on the performance of SVM.
If we normalize our testing data using the mean and variance of training data, F measures are around 20-30\%, while if we normalize testing data using their own mean and variance, F measures can reach almost 80\%.
This may suggest that for SVM, changes relative to long-term trends are more important than absolute values of sensor data.
Therefore, in this paper, we normalize the training and testing datasets for SVM by their own means and variances.
On the other hand, we find that DNN performs better when testing data is normalized using the mean and variance of the training data, and hence normalize it that way.

\section{Implementation and Experimental Setup}\label{sec:implementation}

Our DNNs are implemented using the Chainer deep learning framework~\cite{tokui2015chainer}.
Our SVM is implemented using the scikit-learn machine learning library~\cite{scikit-learn}, which uses libsvm~\cite{Chang2011libsvm} as a backend.

To train a DNN with 100 dimensions of hidden layers, we use a machine equipped with an i7 6700 quad core, DDR4 64 GB RAM, and a NVIDIA GTX 1080 8 GB GPU.
For DNNs with more than 200 dimensions of hidden layers, we use GPU cluster machines equipped with 10 cores of Intel Xeon E5-2630Lv4, 256GB RAM and 8 NVIDIA Tesla P100s.
Training takes about two weeks for the DNN with 100 dimensions of hidden layers with 58 training epochs.
Evaluation of the DNNs is performed on cluster machines equipped with 10 cores of Intel Xeon E5-2630Lv4 and 256GB RAM without using a GPU.
Evaluation takes about 8 hours for the DNN with 100 dimensions of hidden layers.

One-class SVM was trained and tested on a Mac Pro equipped with a 3GHz, 8-core Intel Xeon E5 and 64GB of DDR3 RAM and the aforementioned cluster machines without GPUs.
The running time varied widely by parameters: training took between 11 seconds and 26 hours depending on the parameter settings, while evaluation took up to 7 hours.
The best-performing parameter combination took 30 minutes to train and 10 minutes to evaluate.

\section{Hyper-Parameter Tuning}\label{sec:tune}

Both methods require some tuning of their hyper-parameters, i.e.~parameters whose values are set before learning.
We explain in the following how we performed this.

\subsection{Deep Neural Network}

Our DNN has many parameters, but we tune only the dimensions of intermediate layers (purple in the Fig.~\ref{fig:DNN}) and the threshold value of the outlier factor.
We split the normal log into ten chunks.
Training uses batch learning with a batch size of ten; 100 steps of log entries are learnt at once, and back propagation is unchained after every 100 steps of log entries.
The activation function used is the sigmoid function.
We remark that Rectified Linear Unit (ReLU) was also tried, but training error quickly becomes NaN.
We also tried the architecture in which intermediate layers between output layers and the bilinear functions in the top layers are removed, but we found that the training error does not stabilize with respect to the training epochs.

\begin{figure}[!t]
\centering
\includegraphics[width=\linewidth]{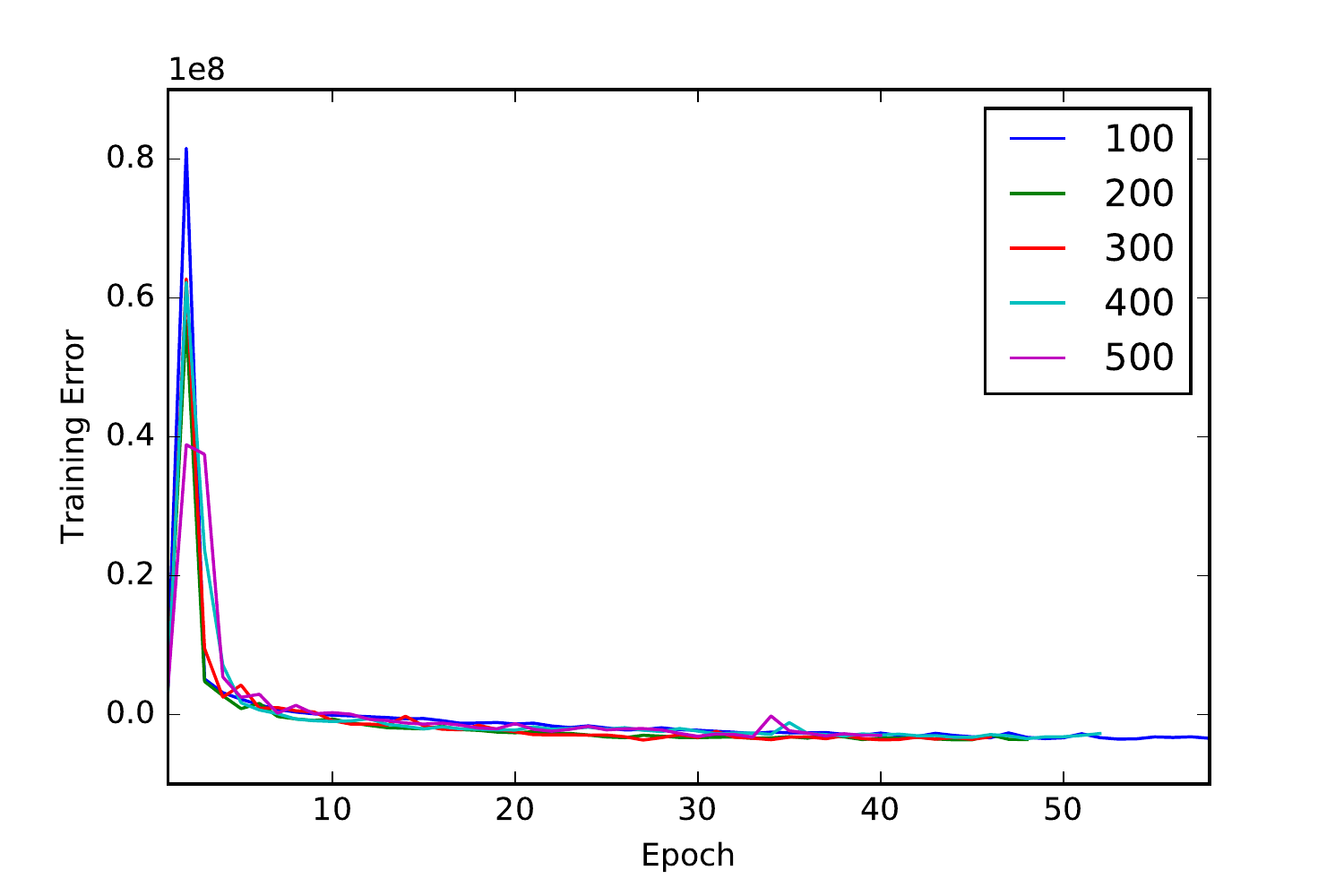}
\caption{Training error of DNNs}
\label{fig:training-error}
\end{figure}

Fig.~\ref{fig:training-error} shows the training error along with the number of training epochs.
We experimented on DNNs with 100--500 dimensions of intermediate layers.
The training error steadily decreases as the number of training epochs is increased, regardless of the dimension of the intermediate layers.

\begin{figure}[!t]
\centering
\includegraphics[width=\linewidth]{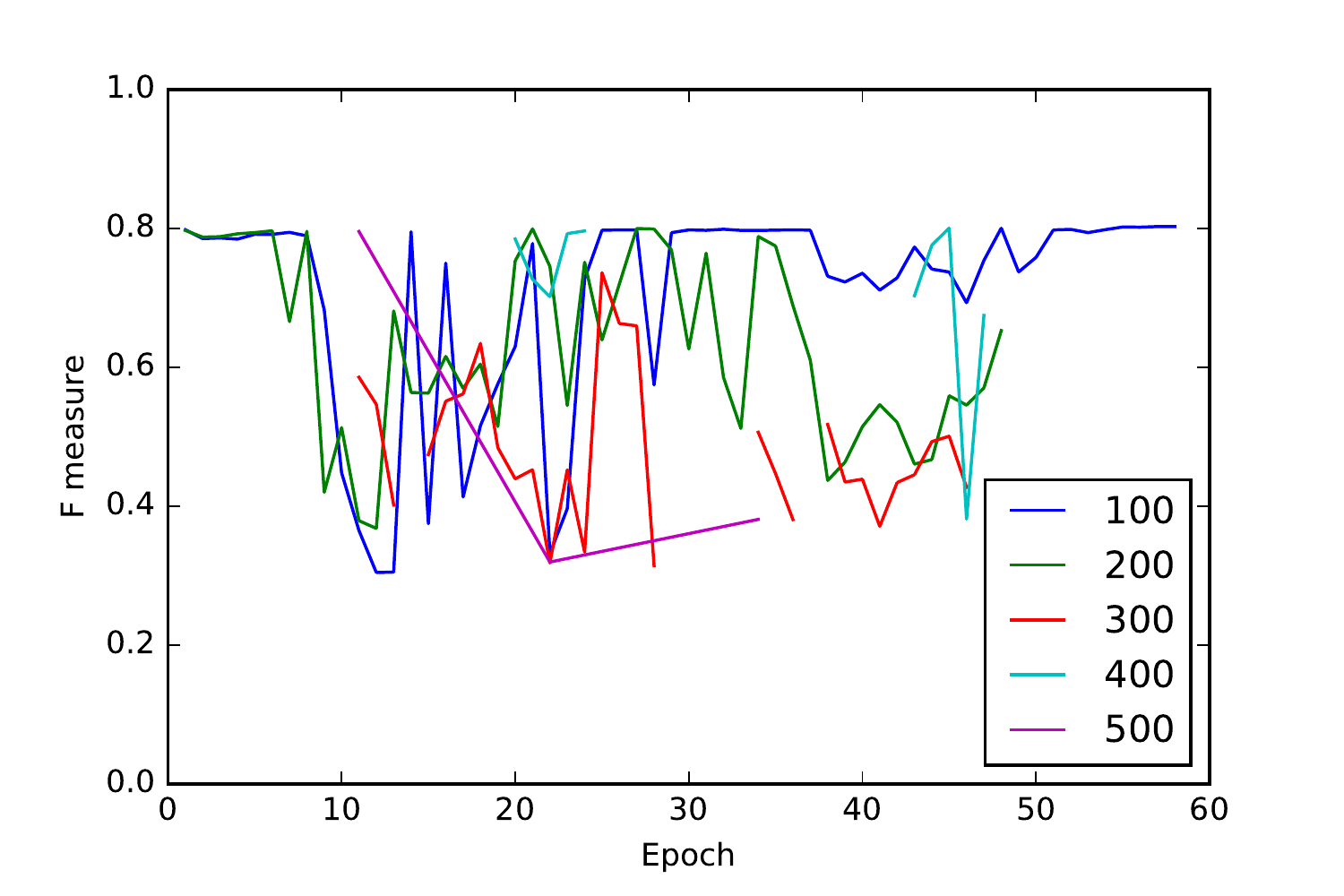}
\caption{F measure}
\label{fig:F-DNN}
\end{figure}

Fig.~\ref{fig:F-DNN} shows the F measure of trained DNNs of each epoch.
Some data points are missing because we could not evaluate all neural models due to limits in our computation budget.
The F measure depends on a threshold value: we maximize the F measure by using a different threshold value for each epoch and DNN.
Interestingly, there is no trend with respect to the number of training epochs.
The first epoch already achieves almost the best F measure, and 10--30 epochs show the worst performance; the Area Under Curve (AUC) has also a similar tendency.
Based on these results, we choose a DNN with 100 dimensions of intermediate layers, which is trained with 58 epochs (the maximal in our experiment) for the evaluation.

By using a sufficient amount of held-out data, we can tune the hyper-parameters without using data that contains anomalies (possibly leading to test and thus generalization errors).
This is desirable, because for most realistic situations, we do not have data with real anomalies, and simulated anomalies may not represent real anomalies.
However, in this experiment, we do not use held-out data to test the accuracy of trained DNNs, because we deem that we do not have enough data.
We use the last day of the normal log as held-out data and test the models.
This method suggests early stopping of training: around 13 epochs.
After 13 epochs, the test error steadily increases.
However, the model obtained with 13 epochs of training underperforms with a wide margin against better trained models  when evaluated with attack data.

\subsection{One-Class SVM}

One-class SVM has three parameters: $w$, $\nu$ and $\gamma$.
As before, $w$ is the size of the sliding window;
$\nu$ is a weight in the range $(0..1]$ that controls the trade-off between mis-classifying normal data as abnormal and the vector-norm of the learned weights (i.e.\ model simplicity);
and $\gamma$ is a coefficient of the kernel.
By default, scikit-learn chooses $\nu = 0.5$ and $\gamma = 1/n$, where $n$ is the number of dimensions in one feature vector (i.e.\ one window).
In our setup, $n = 52w$.

To explore the effects of these parameters at different scales, we first varied the parameters logarithmically (Table~\ref{tab:F-measure-svm}), training and testing one-class SVM with all combinations of $w = 2,4$, $\nu \in \{10^{-4}, 10^{-3}, 10^{-2}, 10^{-1}, 0.5\}$, and $\gamma \in \{10^{-4}, 10^{-3}, 10^{-2}, 10^{-1}, 1\}$.
We varied $w$ through a small range because it directly affects the dimensionality of feature vectors, and high-dimensional feature vectors are known to tend to throw off SVM.
The ranges of $\nu$ and $\gamma$ both contain values in the same ballpark as the defaults in scikit-learn, but as we will now see, those values are suboptimal.

\begin{table}[!t]
\centering
\caption{F measures from logarithmic grid search on $\nu$, and $\gamma$}
\label{tab:F-measure-svm}
\begin{tabular}{lrrrrr}
\multicolumn{6}{c}{$w=2$}\\
\toprule
$\gamma$ $\backslash$ $\nu$& $10^{-4}$& $10^{-3}$& $10^{-2}$& $10^{-1}$& $0.5$\\
$10^{-4}$& 0.02973& 0.08248& 0.12590& 0.47346& 0.31791\\
$10^{-3}$& 0.13399& 0.13924& \textbf{0.77782}& 0.59440& 0.32857\\
$10^{-2}$& 0.69236& 0.68711& 0.63592& 0.49769& 0.29959\\
$10^{-1}$& 0.22105& 0.22103& 0.22149& 0.21845& 0.21471\\
$1.0$& 0.21409& 0.21409& 0.21409& 0.21409& 0.21409\\
\bottomrule
\end{tabular}
\\\vspace{\baselineskip}
\begin{tabular}{lrrrrr}
\multicolumn{6}{c}{$w=4$}\\
\toprule
$\gamma$ $\backslash$ $\nu$& $10^{-4}$& $10^{-3}$& $10^{-2}$& $10^{-1}$& $0.5$\\
$10^{-4}$& 0.05237& 0.08461& 0.12688& 0.50846& 0.32168\\
$10^{-3}$& 0.79140& \textbf{0.79506}& 0.79127& 0.58968& 0.32617\\
$10^{-2}$& 0.53330& 0.53048& 0.49043& 0.37822& 0.26742\\
$10^{-1}$& 0.21452& 0.21452& 0.21451& 0.21435& 0.21433\\
$1.0$& 0.21433& 0.21433& 0.21433& 0.21433& 0.21433\\
\bottomrule
\end{tabular}
\end{table}

The grid search suggests that the best performing instance exists around $w=4$, $\gamma=\nu=10^{-3}$.
We further explore the optimal parameter using random parameter search~\cite{bergstra2012random}.
We fix $w=4$ and generate $\gamma$ and $\nu$ randomly using the exponential distribution scaled by $10^{-3}$.
We test 4204 random instances generated by this method, and improve the F measure to 0.79628.
Analyses in the following section refer to this best-performing instance: $w=4$, $\gamma=0.0008181483058667633$, $\nu=0.004584962079820046$.

\section{Evaluation}\label{sec:evaluation}

\begin{table}[!t]
\centering
\caption{Comparison of DNN and one-class SVM}
\label{tbl:comparison}
\begin{tabular}{lrrr}
\toprule
Method & Precision & Recall & F measure\\
\midrule
DNN & \textbf{0.98295} & 0.67847 & \textbf{0.80281}\\
One Class SVM &0.92500 &\textbf{0.69901} & 0.79628\\
\bottomrule
\end{tabular}
\end{table}

Table \ref{tbl:comparison} presents a comparison of our DNN and one-class SVM on the SWaT attack log, with respect to precision, recall, and F measure of anomalies.
The hyper-parameters were tuned as described in Section~\ref{sec:tune}.
DNN has better precision while SVM has slightly better recall; overall DNN has a slightly better F measure.

It should be noted that false positive and true positive rates, which underlie these statistics, are counted over log entries for DNN whereas they are counted over windows for SVM.
Thus, a direct comparison of the numbers in Table~\ref{tbl:comparison} should only be made with this in mind.
A more in-depth comparison follows, which does confirm the impression given by the table: namely, that both detectors are able to catch anomalies at comparable rates, but SVM is more prone to false alarms.

\begin{figure}[!t]
\centering
\includegraphics[width=\linewidth]{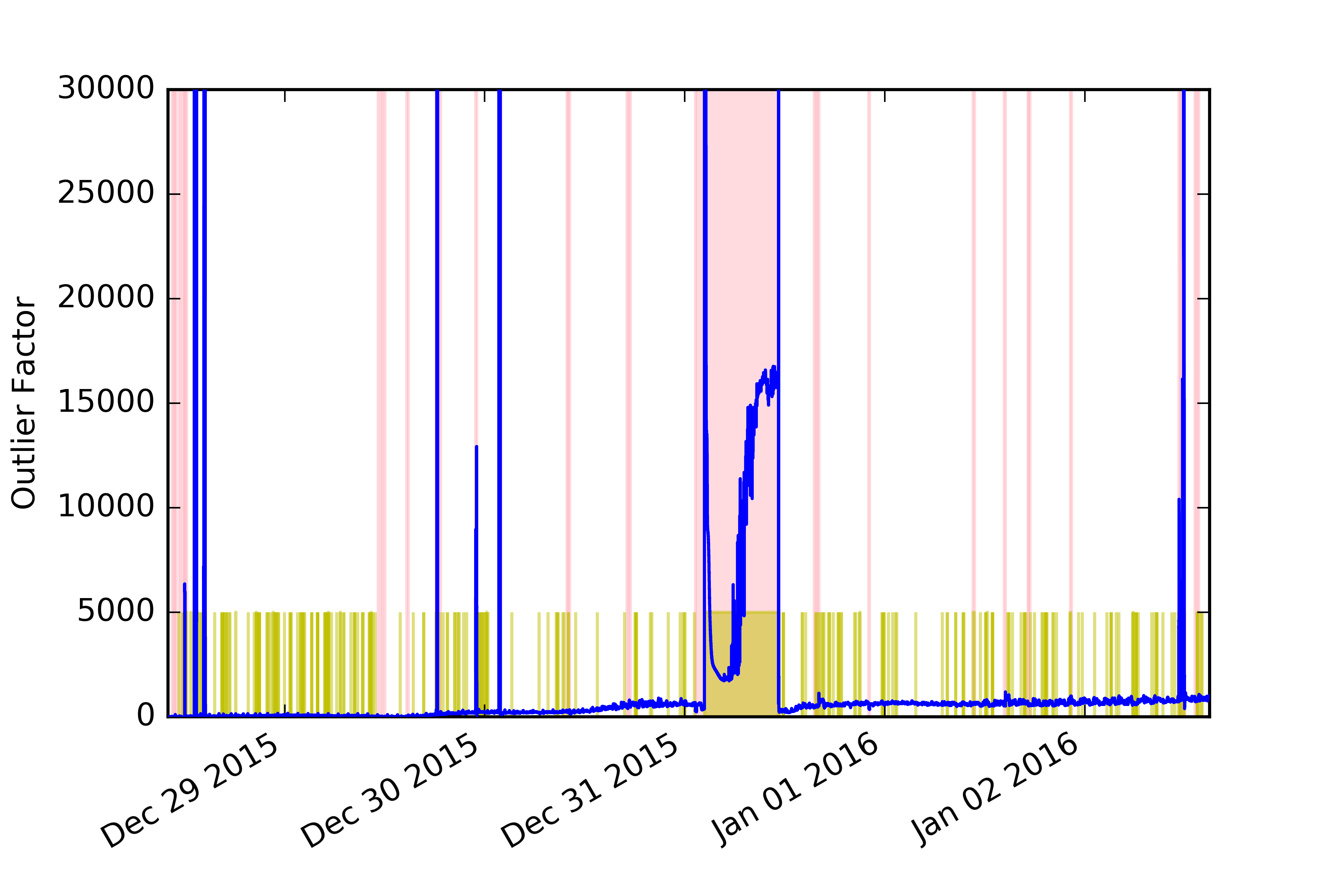}
\caption{Outlier factor and SVM verdicts}
\label{fig:outlier-factor-1}
\end{figure}

Fig.~\ref{fig:outlier-factor-1} depicts how the outlier factor (blue line), SVM verdicts (gold bar indicating an anomaly verdict), and ground truth (pink background indicating an attack) change during the entire attack dataset, spanning from Dec 29.~2015 to Jan 2.~2016.
We can observe that a large outlier factor corresponds to anomalies, while some anomalies do not cause an increase of outlier factor.
SVM emits false alarms intermittently.
Note that at this level of magnification, regions of SVM false alarms appear more densely marked than they actually are.
Overall, SVM emits false alarms on about 0.8\% of non-attack windows.

\begin{figure}[!t]
\centering
\includegraphics[width=\linewidth]{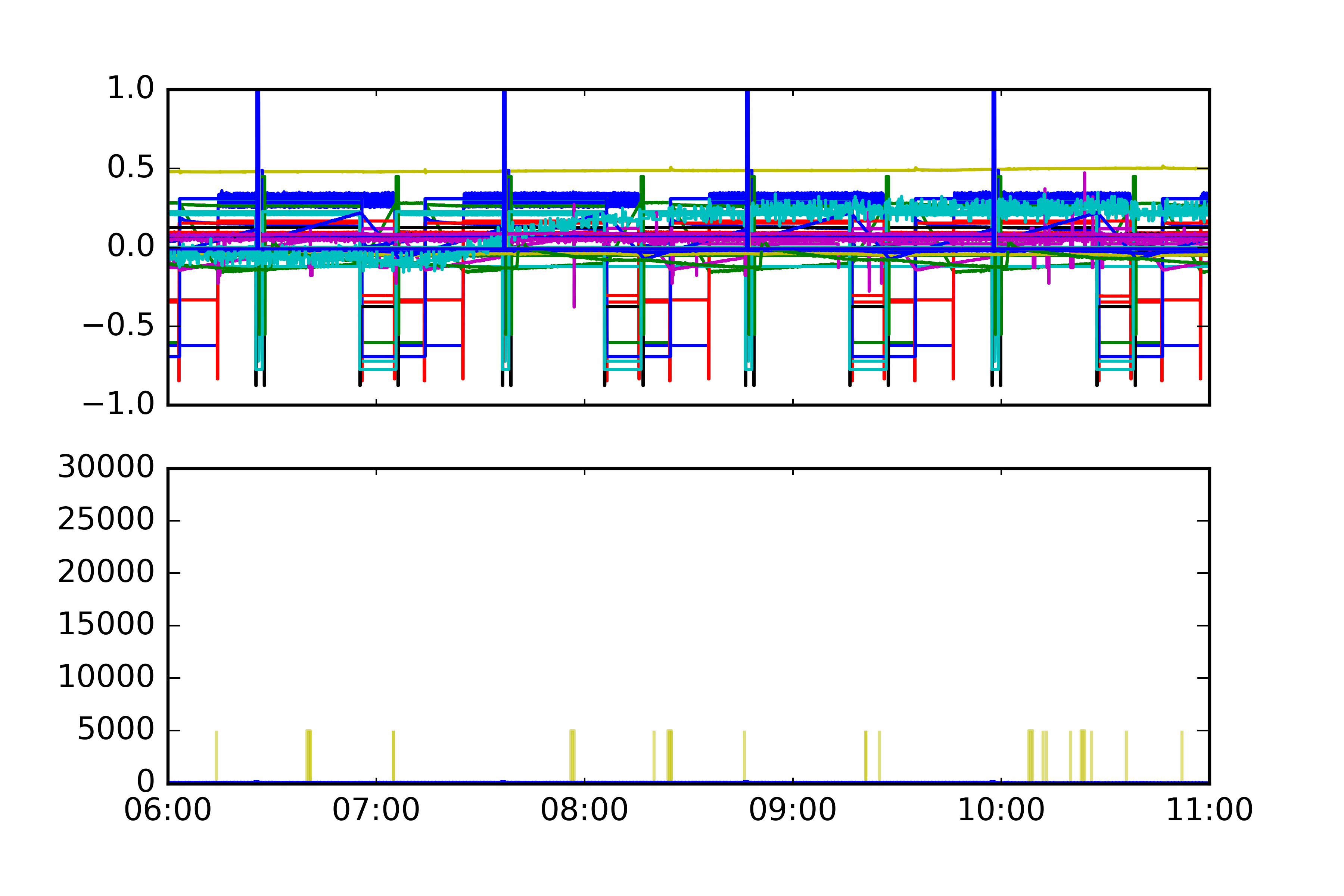}
\caption{SVM false positives (Dec 29.~2015)}
\label{fig:attack-svm-2}
\end{figure}

Fig.~\ref{fig:attack-svm-2} shows some false positives reported by our SVM.
As discussed above, our SVM tends to report false positives intermittently.
The figure suggests that abrupt changes of some sensor values may be the cause.
This is natural because our SVM only uses the values in the moving window, hence longer term trends are not counted at all.

Next, we investigate the effectiveness of the methods at detecting individual attacks.
Table \ref{tbl:attacks} shows the recall rates of both methods for each attack.
The attack IDs and descriptions correspond to those provided in the dataset documentation~\cite{SWaT-Dataset} (note the omission of attacks \#5, \#9, \#12, \#15, \#18, which have no effect on the physical state, and thus no effect on the attack log).
According to this table, if an attack changes sensor values to constant anomalous values, our DNN usually detects it.
On the other hand, our DNN misses the attacks which cause anomalous actuator movements or gradual changes of sensor values.
The behavior of SVM is harder to characterize.
SVM seems more effective at detecting out of range values as in \#7 and \#32, but not always, as in \#31 and \#33.
As shown by \#10 and \#39, SVM sometimes has difficulty in detecting anomalies which are detected by DNN with high precision.

\begin{table*}[!t]
\centering
\caption{Recall rates of DNN and SVM for each attack}
\label{tbl:attacks}
\begin{tabular}{lp{12cm}rr}
\toprule
ID & Description of Attack & DNN & SVM \\
\midrule
1 & Open MV-101 & 0 & 0 \\
2 & Turn on P-102 & 0 & 0 \\
3 & Increase LIT-101 by 1mm every second & 0 & 0 \\
4 & Open MV-504 & 0 & 0.035 \\
6 & Set value of AIT-202  as 6 & 0.717 & 0.720 \\
7 & Water level LIT-301 increased above HH & 0 & 0.888 \\
8 & Set value of DPIT as $>\!\ $40kpa & 0.927 & 0.919 \\
10 & Set value of FIT-401 as $<\!\ $0.7 & 1 & 0.433 \\
11 & Set value of FIT-401 as 0 & 0.978 & 1 \\
13 & Close MV-304 & 0 & 0 \\
14 & Do not let MV-303 open & 0 & 0 \\
16 & Decrease water level LIT-301 by 1mm each second & 0 & 0 \\
17 & Do not let MV-303 open & 0 & 0 \\
19 & Set value of AIT-504 to 16 uS/cm & 0.123 & 0.13 \\
20 & Set value of AIT-504 to 255 uS/cm & 0.845 & 0.848 \\
21 & Keep MV-101 on continuously; Value of LIT-101 set as 700mm & 0 & 0.0167 \\
22 & Stop UV-401; Value of AIT502 set as 150; Force P-501 to remain on & 0.998 & 1 \\
23 & Value of DPIT-301 set to $>\!\ $0.4 bar; Keep MV-302 open; Keep P-602 closed  & 0.876 & 0.875 \\
24 & Turn off P-203 and P-205 & 0 & 0 \\
25 & Set value of LIT-401 as 1000; P402 is kept on & 0 & 0.009 \\
26 & P-101 is turned on continuously; Set value of LIT-301 as 801mm & 0 & 0 \\
27 & Keep P-302 on continuously; Value of  LIT401 set as 600mm till 1:26:01 & 0 & 0 \\
28 & Close P-302 & 0.936 & 0.936 \\
29 & Turn on P-201; Turn on P-203; Turn on P-205 & 0 & 0 \\
30 & Turn P-101 on continuously; Turn MV-101 on continuously; Set value of LIT-101 as 700mm; P-102 started itself because LIT301 level became low & 0 & 0.003 \\
31 & Set LIT-401 to less than L & 0 & 0 \\
32 & Set LIT-301 to above HH & 0 & 0.905 \\
33 & Set LIT-101 to above H & 0 & 0 \\
34 & Turn P-101 off & 0 & 0 \\
35 & Turn P-101 off; Keep P-102 off & 0 & 0 \\
36 & Set LIT-101 to less than LL & 0 & 0.119 \\
37 & Close P-501; Set value of FIT-502 to 1.29  at 11:18:36 & 1 & 1 \\
38 & Set value of AIT402 as 260; Set value of AIT502 to 260 & 0.923 & 0.927 \\
39 & Set value of FIT-401 as 0.5; Set value of AIT-502 as 140 mV & 0.940 & 0 \\
40 & Set value of FIT-401 as 0 & 0.933 & 0.927 \\
41 & Decrease LIT-301 value by 0.5mm per second & 0 & 0.357 \\
\end{tabular}
\end{table*}

Finally, we remark on two threats to the validity of our evaluation:
\begin{enumerate}
  \item Our experiment is limited to the SWaT testbed.
  \item Our experiment is based on deliberately injected anomalies, not anomalies in real life.
\end{enumerate}
Because of (1), it could be possible that our results do not generalize to other CPSs.
Because of (2), it could be possible that our results do not apply to anomalies from real attacks.

\section{Conclusion and Future Work}\label{sec:concl}

In this paper, we investigated the application of unsupervised machine learning to anomaly detection for CPSs.
In particular, we proposed a DNN (adapted to time series data) that implemented a probabilistic outlier detector, and compared its performance against a one-class SVM.
We assessed the two methods in the context of the SWaT testbed~\cite{SWaT-Reference}, a scaled-down but fully operational raw water purification plant.
In particular, we trained on an extensive dataset covering seven days of continuous normal operation, and evaluated the methods using a dataset from four days of attacks~\cite{Goh_et-al17a,SWaT-Dataset}.

We made our comparison based on the precision and recall of detected anomalies in this attack log, finding that the DNN performs slightly better in terms of F measure, with the DNN having better precision and SVM having slightly better recall.
We also found that SVM tended to report false positives intermittently, possibly due to using only the values in a fixed sliding window.
We found that the computation cost for our DNN was much higher: training the DNN took about two weeks, while the best performing SVM needed only about 30 minutes.
The running times for performing anomaly detection on the four days of attack data is also longer for DNN, taking 8 hours for DNN and only about 10 minutes for SVM.
Both methods share some limitations: they both have difficulties in detecting gradual changes of sensor values.
Both methods also have difficulties in detecting anomalous actuator behavior---overcoming this may require taking into account the logic of the controllers.
We plan to tackle these limitations in future work by improving the neural architecture as well as by feature engineering.

The results of our study face two principal threats to validity.
First, we only experimented with a single dataset generated by the SWaT system, meaning that our results may not generalize to other CPSs.
Second, our dataset only contains anomalies arising from deliberately injected attacks; other anomalies may have different characteristics.

Our study can be extended in several directions.
First, we need to improve the performance of our methods further.
In particular, encoding long-term data trends in the feature vectors would improve the performance of both methods, and to be practical, a higher recall rate is necessary.
In addition, we plan to make our detector capable of computing outlier factors for individual sensors and actuators, to infer which sensors or actuators are causing anomalies.
Next, we plan to test our methods more rigorously using
the \emph{SWaT simulator}.
This software faithfully simulates the cyber part of the SWaT testbed, and provides some approximate models for its simpler physical processes (e.g.~water flow).
Using the simulator, we could perform some additional experiments that would otherwise have some safety consequences, e.g.~injecting faulty control software.
Third, we plan to extend our comparisons to other methods beyond DNN and SVM.
In particular, we plan to compare additional neural-based methods~\cite{Malhotra2015,Zhai2016}, and other machine learning and statistical methods.
It would also be interesting to compare specification mining techniques developed in the software engineering field, such as that of Jones et~al.~\cite{Jones2014}.
Finally, we are actively looking for other real-world CPS datasets on which to evaluate our methods.


%
%



\bibliographystyle{IEEEtran}
\bibliography{IEEEabrv,swat-paper}
%
%
%

\end{document}